\pdfoutput=1
\documentclass[11pt]{article}

\usepackage[]{ACL2023}
\usepackage{amsmath} 
\usepackage{amsfonts}
\usepackage{tcolorbox}

\usepackage{libertine}
\usepackage{algorithm}
\usepackage[noend]{algpseudocode}
\usepackage{times}
\usepackage{latexsym}
\usepackage{CJKutf8}
\usepackage{soul}
\usepackage{color, xcolor} 
\definecolor{DarkGreen}{RGB}{0,100,0}
\usepackage{subcaption} 
\usepackage[T1]{fontenc}
\usepackage[utf8]{inputenc}
\usepackage{graphicx}
\usepackage{microtype}

\usepackage{inconsolata}
\usepackage{multirow}

\title{PFID: Privacy First Inference Delegation Framework for LLMs}

\author{%
Haoyan Yang$^{2}$$^\dagger$, Zhitao Li$^{1}$$
^\dagger$, Yong Zhang$^{1}$, Jianzong Wang$^{1}$$^*$, \\ \textbf{Ning Cheng}$^{1}$\textbf{,} \textbf{Ming Li}$^{3}$\textbf{,} \textbf{Jing Xiao}$^{1}$ \\\\
         $^{1}$Ping An Technology (Shenzhen) Co., Ltd., China \\
         $^{2}$New York University \ $^{3}$University of Maryland\\
         \texttt{hy2847@nyu.edu, ztlisz@foxmail.com, jzwang@188.com}
        }

\begin{document}
\maketitle

\def\thefootnote{$\dagger$}\footnotetext{\ These authors contributed equally to this work.}\def\thefootnote{\arabic{footnote}}

\def\thefootnote{*}\footnotetext{\ Corresponding author: Jianzong Wang. }\def\thefootnote{\arabic{footnote}}

\begin{abstract}
This paper introduces a novel privacy-preservation framework named PFID for LLMs that addresses critical privacy concerns by localizing user data through model sharding and singular value decomposition. When users are interacting with LLM systems, their prompts could be subject to being exposed to eavesdroppers within or outside LLM system providers who are interested in collecting users' input. In this work, we proposed a framework to camouflage user input, so as to alleviate privacy issues. Our framework proposes to place model shards on the client and the public server, we sent compressed hidden states instead of prompts to and from servers. Clients have held back information that can re-privatized the hidden states so that overall system performance is comparable to traditional LLMs services. Our framework was designed to be communication efficient, computation can be delegated to the local client so that the server's computation burden can be lightened. We conduct extensive experiments on machine translation tasks to verify our framework's performance. 

\end{abstract}

\section{Introduction}
The widespread use of powerful large language models (LLMs) like insturctGPT \cite{ouyang2022training} and GPT-4 \cite{openai2023gpt4} in various sectors is primarily due to their remarkable capabilities. However, this extensive application brings with it significant concerns regarding user privacy \cite{yao2023survey}. The prevalent method of accessing LLMs through servers or APIs, while convenient, predominantly leads to privacy breach vulnerabilities. This is because user data is stored on servers, resulting in a lack of control over the confidentiality of the data. Addressing this issue is critical, as privacy concerns can lead to user apprehension of personal privacy leak in utilizing these LLMs, thereby undermining their trustworthiness and limiting their potential benefits.

The privacy issues could be more significant in the field of machine translation (MT). People might use cloud base LLMs translation service on daily basis extensively, sensitive personal information like passwords, phone numbers might get capture by eavesdroppers or a honest-but-curious LLMs service providers \cite{lyu-etal-2020-differentially}. Here, we define eavesdroppers as a man-in-the-middle who is interested in peeping user inputs, they could reside illegally within service providers or intercept user transmission to and from service providers over the internet.

\begin{figure}[t]
  \centering
  \small
  \includegraphics[width=0.48\textwidth]{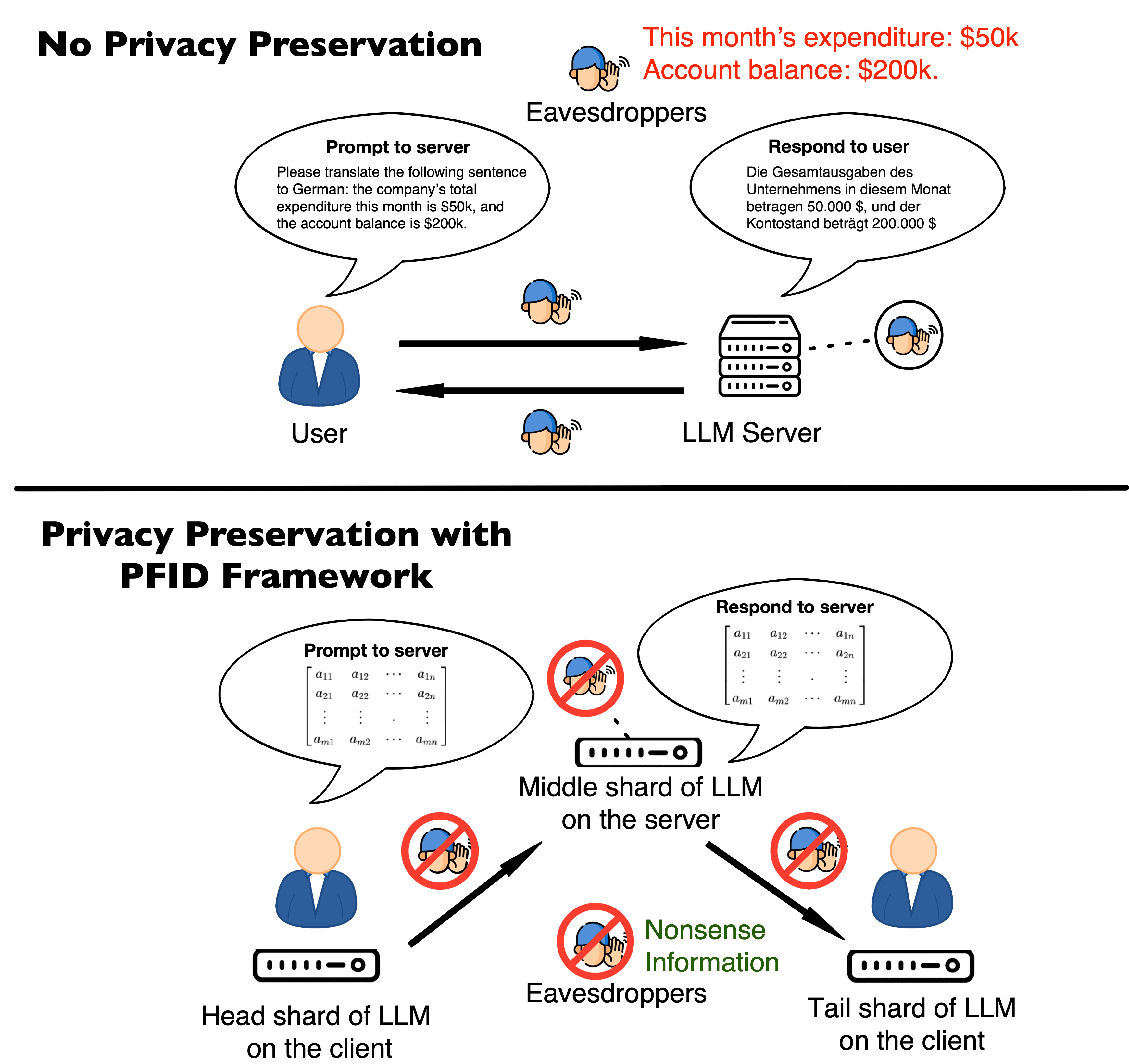}
  \caption{There are two scenarios for interacting with a LLM server: without privacy protection, private data is vulnerable to eavesdroppers; with the PFID Framework, data is encrypted, preventing eavesdroppers from accessing sensitive information.}
  \label{fig-sub}
\end{figure}

To address above mention problems, as shown in Figure~\ref{fig-sub}, we proposes a new privacy first inference delegation (PFID) framework for LLMs, which localizes user data on client device through model sharding and data truncation, safeguarding users' privacy. We propose to split LLMs into three shards, head, middle and tail. Head and tail segments were put on client local device, middle segment was proprietary and resides on server only. Only hidden states computed by the shards were communicated between server and local client instead of prompts. We attempt a compromise between open-source and proprietary models for LLMs. For proprietary LLMs, businesses can retain the core parts of the model on their servers while distributing parts of the model to users. This approach can reduce GPU usage on the server side and increase user engagement, generating economic benefits. For users, running a giant open-source model locally is impractical. In this context, using the PFID framework allows each person to only run a part of the model, enabling the widespread use of super-large models without compromising privacy. Importantly, this is the first known split-and-play, general privacy-preserving inference paradigm for LLMs. The PFID framework aims to establish a distributed inference framework that allows everyone to use LLMs in a personalized manner with enhanced computational resources while maintaining privacy, benefiting both users and businesses. Our main contributions are as follow:

\begin{itemize}
   \item Our research introduces a novel inference framework for model sharding within LLMs that focuses on preserving privacy while distributing the computational workload of autoregressive tasks. Our framework is designed to be split-and-play, no training is needed. 
   \item We develop a mechanism termed ‘re-privatization’ that enables normal auto-decoding process while protecting user privacy. This re-privatization methodology shed lights on previously might overlooked stratified information contains in different collections of singular values and vectors of hidden states.
   \item To facilitate both communication efficiency and secure confinement of private information, we propose the adoption of truncated singular value decomposition techniques. These techniques compress the hidden state representations within LLMs, striking a balance between the reduction of data transmission overhead and the retention of sensitive information.
 \end{itemize}

\section{Related Work}
LLMs show promising performance in NMT \cite{jiao2023chatgpt} \cite{hendy2023good}. A few researchers focus on the privacy aspects of applying such approach, initial approaches to privacy protection involved fine-tuning the models using techniques like differentially private fine-tuning \cite{yu2022differentially} or prompt tuning \cite{li2023privacypreserving, hong2023dpopt}. However, these methods are costly and prompts may not adapt well to all LLMs. 

Another approach is to protect privacy at the input level by masking sensitive entities, as explored in \cite{chen2023hide} and \cite{kan2023protecting}. Yet, in the age of advanced LLMs, simply masking private entities may not suffice due to the models' sophisticated reasoning capabilities and the risk of context-based information leakage.

To address these challenges, a novel paradigm has been proposed where computations sensitive to privacy are distributed on local devices, and shared computations are handled in the cloud \cite{wang2023privatelora}. Despite this innovation, the transmitted embeddings could potentially be decoded, which does not guarantee strict privacy protection. To overcome this, \cite{mai2023splitanddenoise} adds noise to avoid the decoding issue but fails to address the high data transmission costs between local and cloud servers. Another line of research focused on perturbing the original input \cite{lin2024promptcrypt} so that ture input is masked.

\section{Methodology}

\begin{figure*}[h]
  \centering
  \small
  \includegraphics[width=1\textwidth]{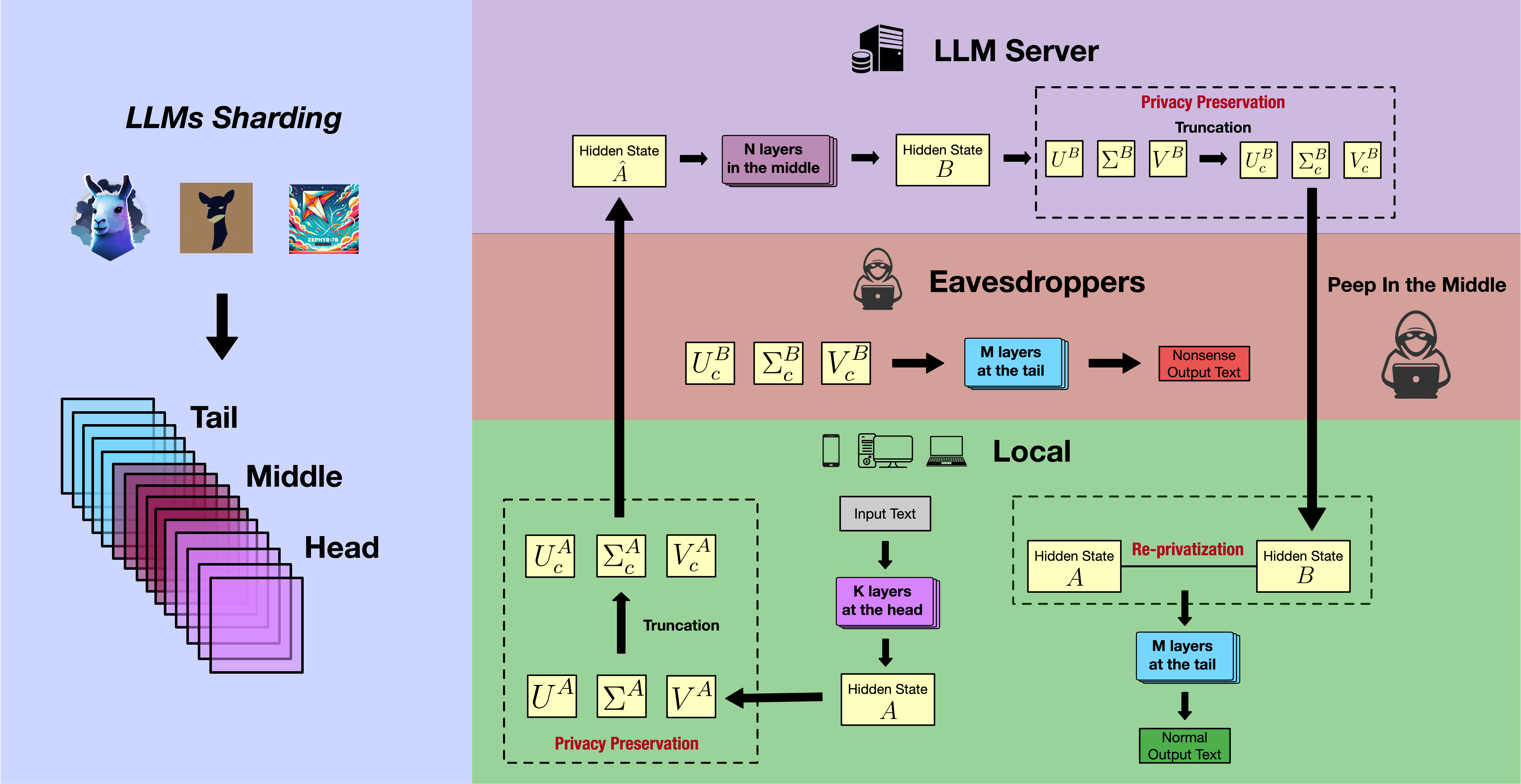}
  \caption{The proposed methods comprise three parts: the head and tail on the client side, potentially open-source, and the proprietary middle on the server. The head processes user input into a hidden state matrix, reduced via truncated SVD to keep the top K components, and sent to the server. The server further process and then compresses the matrix, which the tail then reconstructs for the final output. Results may differ between honest-but-curious servers and direct local processing.}
  \label{fig-main}
\end{figure*}

\subsection{Problem Definition}

The proposed framework encompasses a tripartite structure consisting of a local client, an adversarial eavesdroppers, and the language service provider as illustrated in Figure \ref{fig-main}. The adversarial entity exhibits capability to engage in eavesdropping and interception of the data transmission between the client and the server. The objective of the adversarial entity is to ascertain the content of the client's input prompt and the subsequent tokens generated by the service. Furthermore, the architecture delineates that the head and tail segments of the system are released under an open source license, ensuring that decoding operations with hidden states should be executable by all three aforementioned parties.

Let $M$ be a Large Language Model (LLM) shared by a local client $l$ and eavesdroppers $A$. Given an input token $x \in \mathcal{X}$, where $\mathcal{X}$ is the set of all possible input tokens, we observe the output of the model as $M_l(x)$ when used by the local client and $M_A(x)$ when used by the eavesdropper.

Assuming identical initial states, environments, and inputs for both the local client and eavesdropper, define the conditions under which the following holds:
\begin{equation}
M_L(x) \neq M_A(x)
\end{equation}

We quantify the effect of privacy preserving by measuring the evaluation metric gap between the generated tokens for the client and the eavesdropper, respectively.

\subsection{Compressed Sharding}

In the era of LLMs, models are large and gpu cost for computation is high. Meanwhile, Personal computation power are growing fast, people can now run multi-billion transformer model on smartphones, regular household PCs can run even larger model. It seems like a straightforward idea to delegate some part of large model to client side to cut down GPU usage and increase throughput.

However, the network IO cost for transmitting hidden states to and from server in an auto-regressive manner is prohibitively high. Hence, we develop an truncation approach to cut down the network IO cost.

\subsubsection{Model Sharding}

Transformers layers \cite{NIPS2017_3f5ee243} processes sequential input data in an non-recurrently fashion, such as GPT (Generative Pre-training Transformer) and its derivatives, are fundamentally auto-regressive in nature. This section focus on the decoder-only transformers. This intrinsic characteristic can be encapsulated by the symbol \(\pi\left(x_t|x_{<t}\right)\), which denotes the probability of the model generating a token \(x_t\) given all the previous tokens \(x_{<t}\) in a sequence. In essence, these models iteratively predict the next word or token by conditioning the probability distribution on the sequence of tokens that has been generated thus far. The auto-regressive property ensures that LLMs generate coherent and contextually relevant text, as each step in the generation process takes into account the entire history of the sequence.

Let $X^{(l)}$ be the input to the $l^{th}$ layer of the transformer. The hidden state $\mathbf{H^{(l)}}$ is obtained by applying the self-attention mechanism:

\begin{equation}
\mathbf{H^{(l)}} = \text{DecoderLayer}(Q^{(l)}, K^{(l)}, V^{(l)})
\end{equation}

\begin{equation}
X^{(l+1)} = \text{FeedForward}(\mathbf{H^{(l)}})
\end{equation}

where $Q^{(l)}$, $K^{(l)}$, and $V^{(l)}$ represent the queries, keys, and values, which are linear projections of the input $X^{(l)}$. And the output of the self-attention mechanism is then passed through a feed-forward neural network to produce the final hidden state $\mathbf{H^{(l)}}$ which is then used as the input to the next layer. It is possible to split a model into any number of segments as long as hidden states were passed between them. For a decoder model with $N$ layers. We empirically divide decoder-only transformer into \textbf{Head}, \textbf{Middle} and \textbf{Tail} segments, each segment comprises a number of successive Transformer layers. Notably, this division requires no training.

The \textbf{Head} consists of the initial $N_h$ layers of the model. This segment primarily focuses on early contextual processing of the input sequence. The \textbf{Middle} segment contains a series of layers that follow the head and continue to develop deeper representations, which consist of layers $N_h+1$ to $N_h + N_m$ with each layer taking $\mathbf{H}^{(N_h)}_{\text{head}}$ as input. Finally, the \textbf{Tail} is the concluding segment of the model, composed of the remaining $N-N_h-N_m$ layers taking $\mathbf{H}^{(N_m)}_{\text{middle}}$ as input. It completes the transformation of representations into output token predictions. $h$ and $m$ are hyper parameters, which should be tuned carefully.

\subsubsection{Hidden State Truncation}
Truncated Singular Value Decomposition (SVD) is a technique that approximates a matrix by retaining only the most significant singular values and discarding the smaller ones, which often correspond to noise or less important information. Given a hidden state matrix $\mathbf{\mathbf{H^{(l)}}} \in \mathbb{R}^{d \times n}$ from the $l$ layer of a decoder-only transformer, truncated SVD decomposes $\mathbf{H}$ into:

\begin{equation}
\mathbf{H} = \mathbf{U}_k\mathbf{\Sigma}_k\mathbf{V}_k^*
\end{equation}

Here, $\mathbf{U}_k \in \mathbb{R}^{d \times k}$ contains the first $k$ left singular vectors, $\mathbf{\Sigma}_k \in \mathbb{R}^{k \times k}$ is a diagonal matrix with the top $k$ singular values, and $\mathbf{V}_k^* \in \mathbb{R}^{k \times n}$ contains the first $k$ right singular vectors.

By keeping only the top $p\%$ dimension these singular values, we can create an approximation $\hat{\mathbf{H}}$ of the original hidden state, capturing the significant enough components. We empirically find out that the reduction process effectively filters out the less significant singular values, which often correspond to finer details and noise in the hidden states, and usually about privacy.  The self-supervised training process of LLMs force model to recognise main ideas so as to achieve lower overall loss and hence the top singular values should corresponding to those information. Although cutting of singular values may result in loss of specific information, the preserved structure can be sufficient for capturing the overall patterns and relationships present in the original data.

\subsubsection{Re-privatization}

Modern LLM service employ a text-to-text interface, end user upload plain text to centralized service, but transmitting text to public server can not guarantee privacy. We proposed to send hidden states instead so that public server can not know what input text but client can still utilize centralized LLM service.  Data privatization is done locally on client side, only truncated hidden states are send to public server so that ear-dropping eavesdropper can not know exact hidden states value so as to figure out what input is.

The key for our framework to work is that client have keep-back information from server, private hidden states to enable normal auto-regressive generation, while result generated by the same tail segments of model on server will be different or even become gibberish due to client's hidden state truncation process. 
We call this process re-privatization, meaning to infuse public hidden states with local held back information. Formula for re-privatization is as follow. 

\begin{equation}
\mathbf{H'} = \mathbf{\hat{H}}^{(N_m)}_{\text{middle}} + \Omega * \mathbf{H}^{(N_h)}_{\text{head}}
\end{equation}

where $\mathbf{H'}$ denotes the updated hidden state after integrating the truncated SVD components and additional computations, and was then feed in to the tail segments. $\Omega$ is a hyper paramater that control the level of residual connection. $\mathbf{\hat{H}}^{(N_m)}_{\text{middle}}$ denotes the approximation of $\mathbf{H}^{(N_m)}_{\text{middle}}$ with top $k_t$ singular values. Note that 
$\mathbf{H}^{(N_h)}_{\text{head}}$ is at client local memory, so no approximation is needed. 

By keeping only the top $k$ singular values, we focus on the most informative parts of the hidden state while details are omited. The subsequent layers of computation on $\mathbf{H}_k$ may be designed to perform certain transformations and learn representations that are important for the task at hand. Adding the result of these computations back to the original hidden state $\mathbf{H}$ can be seen as a form of residual connection or skip connection. Our re-privatization features a scalar hyper paramaters $\Omega$ that controls degree of residual fusion, it is not a tune-able parameter. The loss of information due to SVD can help keep pravicy info from eavesdropper decoding while local re-privatization compensate for the details loss.

\begin{algorithm}{h}
\caption{New Token Generation with PFID on Client Side}
\begin{algorithmic}[1]

\State $head \gets \text{Initialize head segs of LLM}$
\State $mid \gets \text{Initialize middle segs of LLM}$
\State $tail \gets \text{Initialize tail segs of LLM}$
\State \textbf{Input}: prompts
\State \textbf{Output}: Predicted Next Token

\Function{ClientGenNewToken}{prompts}

    \State $headOutput \gets head(\text{prompts})$

    \State $U,S,V \gets \text{TruncatedSVD}(headOutput)$ \Comment{Select top $k_h$ singular values}

    \State $midInput \gets  U[:,1:k_h] \cdot \text{diag}(topKValues) \cdot V[:,1:k_h]^T$
    \State $midOutput \gets mid(midInput)$

    \State $U',S',V' \gets \text{TruncatedSVD}(midOutput)$ \Comment{Select top $k_t$ singular values}
    \State $tailInput \gets U'[:,1:k_t] \cdot \text{diag}(S'[1:k_t]) \cdot V'[:,1:k_t]^T$

    \State $tailOutput \gets tail(tailInput + \Omega * headOutput)$
    \State $NextToken \gets LlmHead(tailOutput)$
    \State \Return $NextToken$
\EndFunction
\label{algo1}
\end{algorithmic}
\end{algorithm}

\section{Experiments}

\subsection {Eavesdropping}
In our experiment, we simulated potential privacy eavesdropping within the PFID, specifically the eavesdroppers decode the hidden states transmitted from the public server back to the local. To assess the robustness of our framework in protecting privacy, we compared the decoding results of the \textbf{eavesdroppers} with those of the \textbf{local}. Additionally, we compared the decoding results of the local within the PFID to the inference results of original \textbf{pipeline}. The two comparisons were conducted to verify that PFID can preserve privacy completely without significant loss of information.

\subsection {Experiment Setup}
\subsubsection{Baseline Models}
We employed three of the current mainstream LLMs as baseline models for our experiments: Vicuna-7B \cite{vicuna2023}, Zephyr-7B \cite{tunstall2023zephyr} and Llama-7B \cite{touvron2023llama}. This selection aims to demonstrate the broad applicability of the PFID. All above mentioned models has 32 layers. 

\subsubsection{Datasets}
We tested the PFID on the WMT23 \footnote{WMT23 data is from \url{https://www2.statmt.org/wmt23/mtdata/}} dataset, conducting experiments on translations from six different languages into English. The translation task involves various privacies and naturally fits by allowing the comparison of the translated text with standard results.

\subsubsection{Metrics}
We evaluated the model's performance using two metrics: BLEU \cite{papineni2002bleu}, which checks word precision and COMET \cite{rei2020comet}, which assesses semantic and contextual alignment. 
\subsubsection{Inference Parameter Setting}
During the inference process, the parameters we set are as follows:

\begin{table}[ht]
\centering
\small
\caption{Parameters used in the inference process.}
\begin{tabular}{cc}
\hline
\textbf{Parameter} & \textbf{Value} \\
\hline
Temperature & 0.7 \\
Top-p  & 0.5 \\
Top-k  & 50 \\
Maximum new tokens & 300 \\
\hline
\end{tabular}
\label{tab1}
\end{table}

\subsection{PFID Hyperparameter Tuning}
\renewcommand{\labelitemi}{}
In the PFID, there are four important hyperparameters: \\
\textbf{Layer Range: } presented in the form of $(K, N)$, it represents dividing into the first $K$ layers, from $K$ to $N$, and from $N$ to the last three segments. \\
\textbf{Omega ($\Omega$)}: the weight of incorporating original local information when the hidden state is passed back locally. \\
\textbf{Phead}: the truncation ratio when transmitting from local to the public server, with a higher value indicating more truncation. \\
\textbf{Ptail}: the truncation ratio when transmitting back from the public server to local.

\begin{figure}
    \centering
    \begin{subfigure}[b]{0.23\textwidth}
        \includegraphics[width=\textwidth]{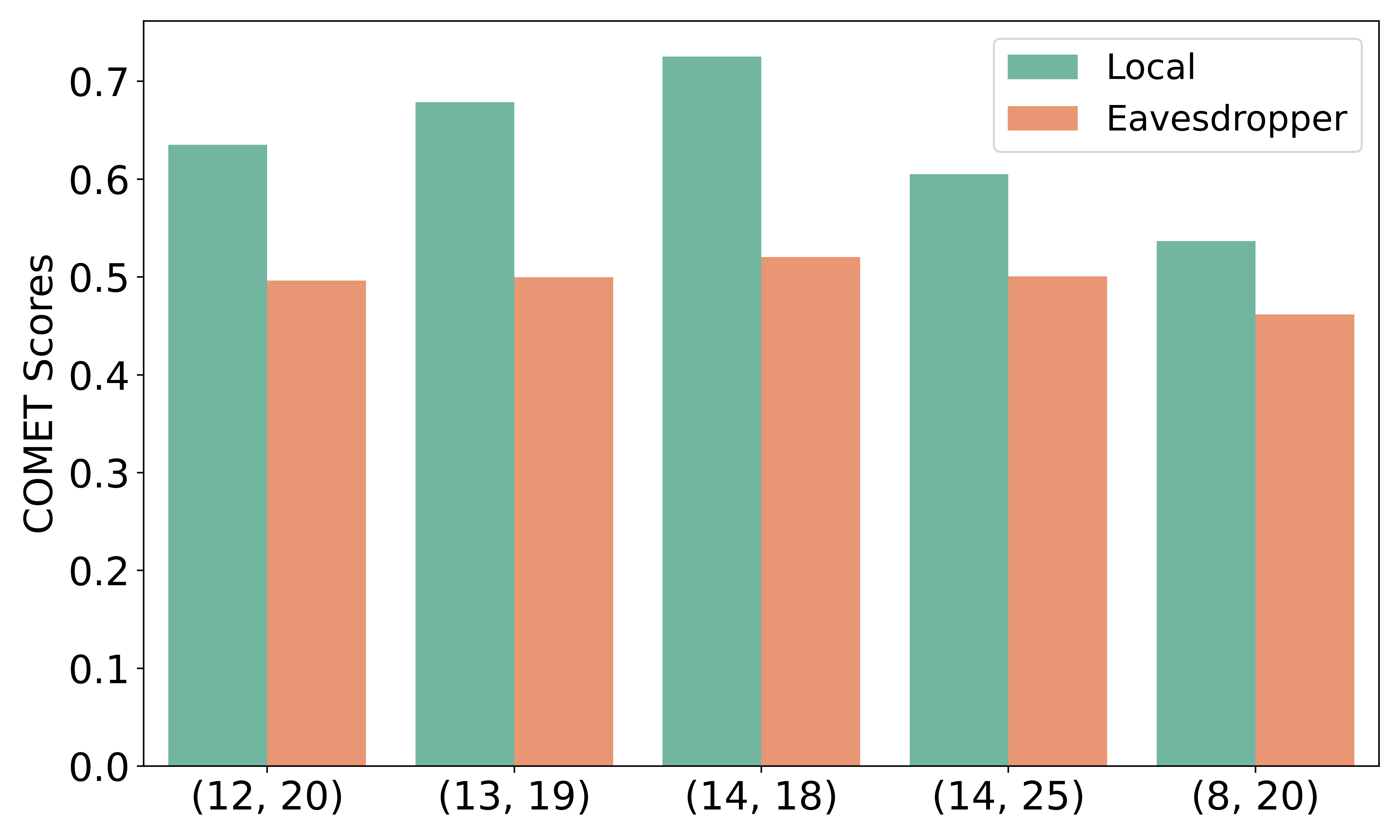}
        \caption{Layer Range}
        \label{fig2:sub1}
    \end{subfigure}
    \hfill 
    \begin{subfigure}[b]{0.23\textwidth}
        \includegraphics[width=\textwidth]{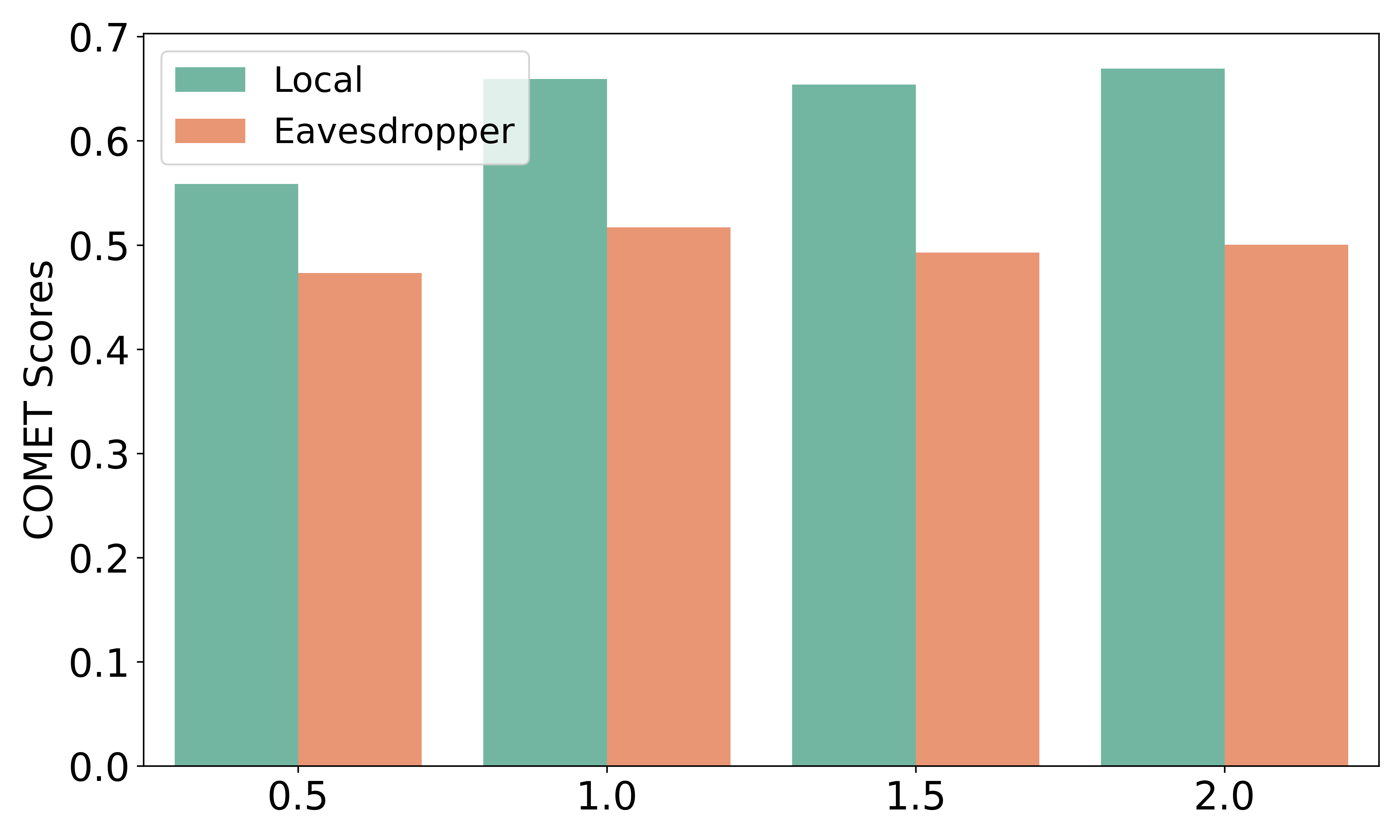}
        \caption{Omega}
        \label{fig2:sub2}
    \end{subfigure}
    \begin{subfigure}[b]{0.23\textwidth}
        \includegraphics[width=\textwidth]{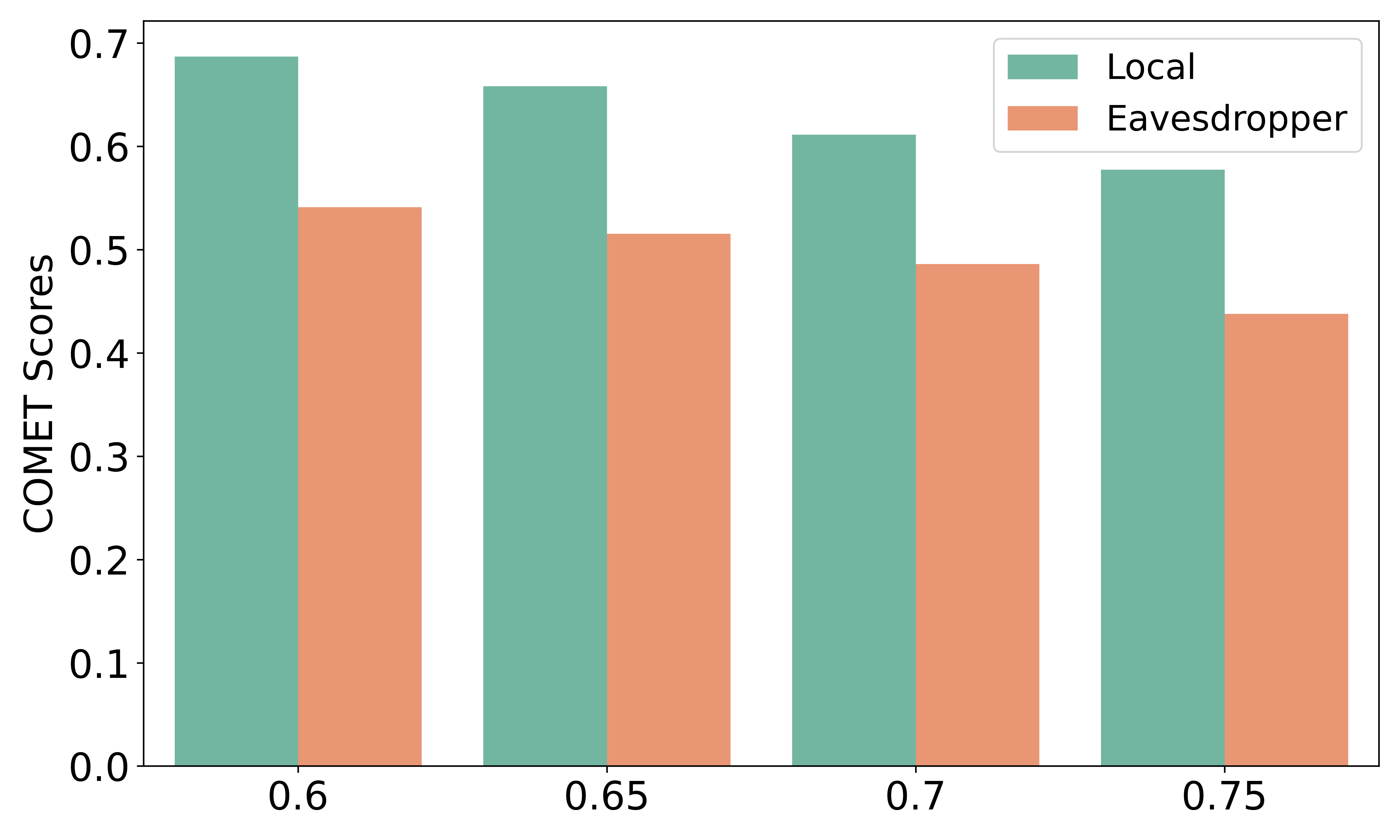}
        \caption{Phead}
        \label{fig2:sub3}
    \end{subfigure}
    \hfill
    \begin{subfigure}[b]{0.23\textwidth}
        \includegraphics[width=\textwidth]{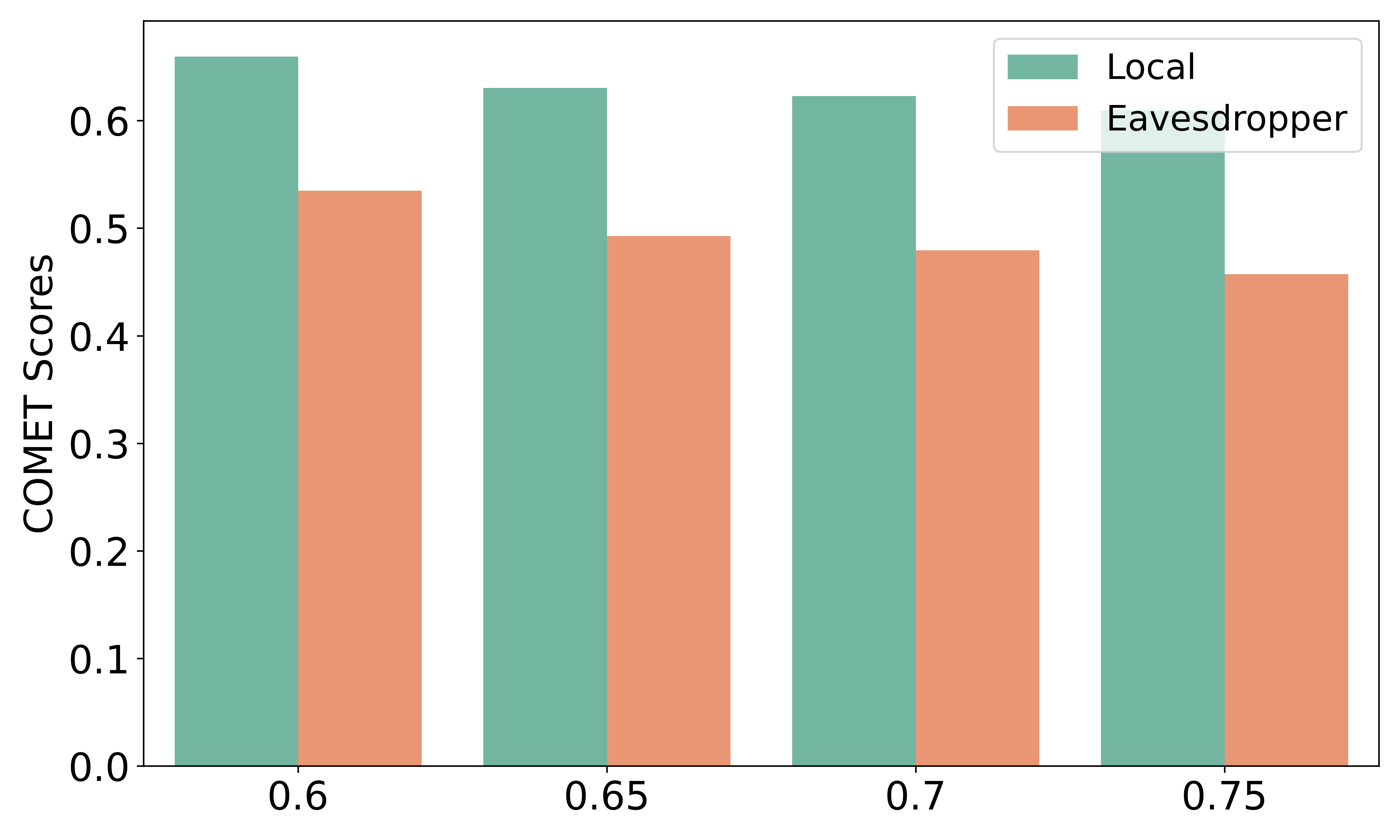}
        \caption{Ptail}
        \label{fig2:sub4}
    \end{subfigure}

    \caption{Optimal PFID hyperparameters for the WMT23 dataset, showcasing Layer Range, Omega, Phead, and Ptail tuning results.}
    \label{fig2}
\end{figure}

As shown in Figure~\ref{fig2}, after conducting hyperparameter tuning experiments on the WMT23 dataset using Vicuna, we identified a set of parameters that resulted in great PFID performance, specifically in the following table:

\begin{table}[ht]
\centering
\small
\caption{Optimal hyperparameters used in the PFID.}
\begin{tabular}{cc}
\hline
\textbf{Hyperparameter} & \textbf{Value} \\
\hline
Layer Range & (13,19) / (14,18) \\
Omega  & 1 \\
Phead  & 0.65 / 0.7 \\
Ptail & 0.75 \\
\hline
\end{tabular}
\label{tab2}
\end{table}

\subsection{Comparative Results}
We compared the performance of three models in terms of local and eavesdropper decoding under both the original pipeline and the PFID framework, as shown in Table~\ref{tab3}. The results indicate that the performance of local decoding is consistently higher than that of eavesdropper decoding across all datasets, suggesting that privacy is well protected under the PFID framework. Furthermore, the performance of local decoding is comparable to that achieved with the original pipeline, indicating that the truncation of data by PFID does not lead to significant information loss. The ability of local decoding to effectively reconstruct information, in contrast to the inability of eavesdroppers, validates the PFID framework's capability to safeguard privacy effectively.

\begin{table*}[t]
\centering
\small
\caption{Comparative results various xx-en language pairs for the Vicuna-7B, Zephyr-7B-beta, and Llama-2-7B-chat models, comparing pipeline, local, and eavesdropper.}
\begin{tabular}{ccccccccc}
\hline
\multicolumn{1}{l}{}             &                        &          & zh-en & ja-en & de-en & uk-en & ru-en & he-en \\
\hline
\multirow{6}{*}{Vicuna-7B}       & \multirow{3}{*}{COMET}  & Pipeline & 0.771      &   0.758    &   0.820    &  0.843     & 0.782      &  0.670     \\  
                                 &                        & Local    &  0.724     &    0.739   &  0.808     &  0.826     & 0.756       & 0.629       \\
                                 &                        & Eavesdropper &  0.502     &  0.443      & 0.585     &  0.610    &  0.519      & 0.478        \\
\cline{2-9}
                                 & \multirow{3}{*}{BLEU} & Pipeline &  20.20     &   11.71    &   35.12    &   38.52    &  19.08     &   20.23     \\  
                                 &                        & Local    & 15.45      & 10.86     &  29.57      &  37.13     & 18.99      & 18.09      \\
                                 &                        & Eavesdropper &  8.11      &  3.52     &  21.42     &  17.73      & 7.28      & 10.37      \\
                                 \hline
\multirow{6}{*}{Zephyr-7B-beta}  & \multirow{3}{*}{COMET}  & Pipeline &  0.739     &   0.695    &   0.776    &  0.749     &  0.728     &   0.608    \\
                                 &                        & Local    &  0.733     &  0.700     &  0.763     &  0.796     &  0.759     &  0.607     \\
                                 &                        & Eavesdropper & 0.537     &  0.489     &   0.551    &  0.568     &  0.532     &   0.473   \\
\cline{2-9}
                                 & \multirow{3}{*}{BLEU} & Pipeline &  14.50     &  7.01     &  26.41     &   22.28    &   16.27    &  11.57     \\
                                 &                        & Local    & 12.75      &  6.17     &   21.99    & 22.47      & 18.38      &   8.45    \\
                                 &                        & Eavesdropper & 6.84      &  2.32     &  14.74     &  11.85     &   7.90    &   5.03    \\
                                 \hline
\multirow{6}{*}{Llama-2-7B-chat} & \multirow{3}{*}{COMET}  & Pipeline &  0.616     &  0.592     &  0.651     &  0.575     &  0.706     &  0.453     \\
                                 &                        & Local    &  0.611     &  0.576     &  0.585     &  0.610     &  0.642     &   0.494    \\
                                 &                        & Eavesdropper &   0.463    &  0.423     &   0.428    &   0.443    &   0.472    &   0.413    \\
\cline{2-9}
                                 & \multirow{3}{*}{BLEU} & Pipeline &  7.84     &   4.67    &  13.11     &  8.49     &   17.53    &  1.51     \\
                                 &                        & Local    &  7.14     &  3.02     &  11.99    &  9.14     &   14.01    &   2.04    \\
                                 &                        & Eavesdropper &   4.54    &  1.31     &  7.54     &  5.05     &    9.55   & 1.00  \\
\hline
\end{tabular}
\label{tab3}
\end{table*}

\subsection{Framework Complexity}
In this section, we analysis the computational complexity and communication complexity introduced by the PFID to both client side and server side.

\begin{figure}[t]
% % https://colab.research.google.com/drive/1EG82reDnLoigGXugnwAZsoBVfbi9gu87?usp=sharing
\includegraphics[width=0.4\textwidth]{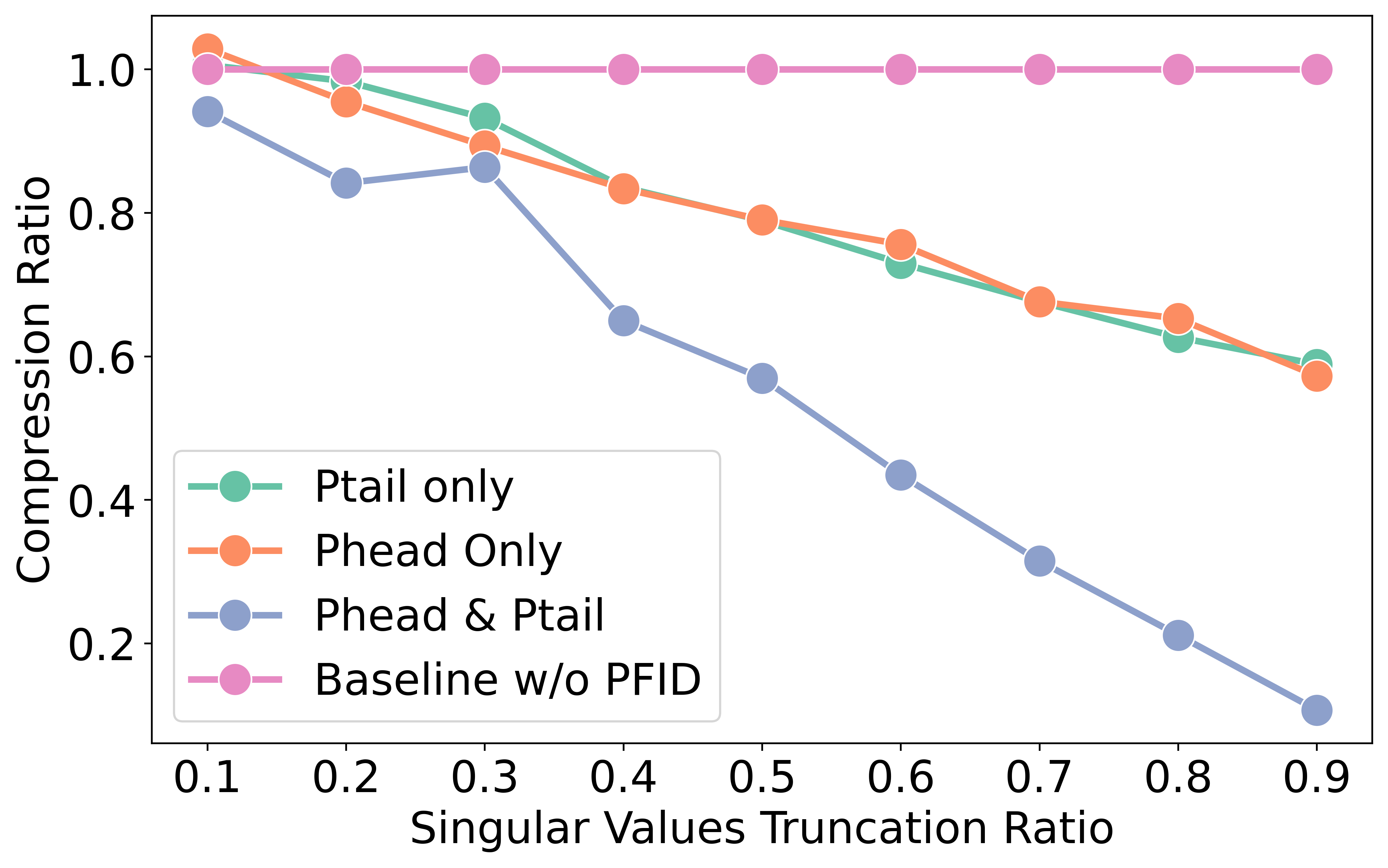}
\centering
\caption{The graph shows the PFID's truncation ratio in a linear fashion with data compression ratio. It measures amnout of data sent to/from the server with and without PFID during performing task. "Ptail" represents discarding singular values between middle and tail segments, "Phead" between the first and middle, and "Phead \& Ptail" discards across all segments.}
\label{fig4}
\end{figure}

\textbf{Communication complexity} Communication complexity is composed of following cost for each token generation. (1) client upload truncated matrix ($U$,$V$,$\Sigma$) to server. (2) server download truncated matrix ($U$,$V$,$\Sigma$) to client. Figure~\ref{fig4} showcases our truncation results. With PFID, the communication budget decrease linearly in accordance with the ratio of singular values truncation.

\textbf{Computation complexity} Both server and user computation complexity can be broke down into (1) svd computation on the client side, (2) svd computation on the server side. We use truncated SVD with randomized SVD, hence the computational complexity is $O(k^2dp)$. In order to avoid Truncated SVD deteriorate into full SVD in terms of computational complexity, we are selecting singulars values by index instead of filtering on cumulative percentage. 

\subsection{Ablation Study}
\begin{figure}[ht]
    \centering
    \includegraphics[width=0.4\textwidth]{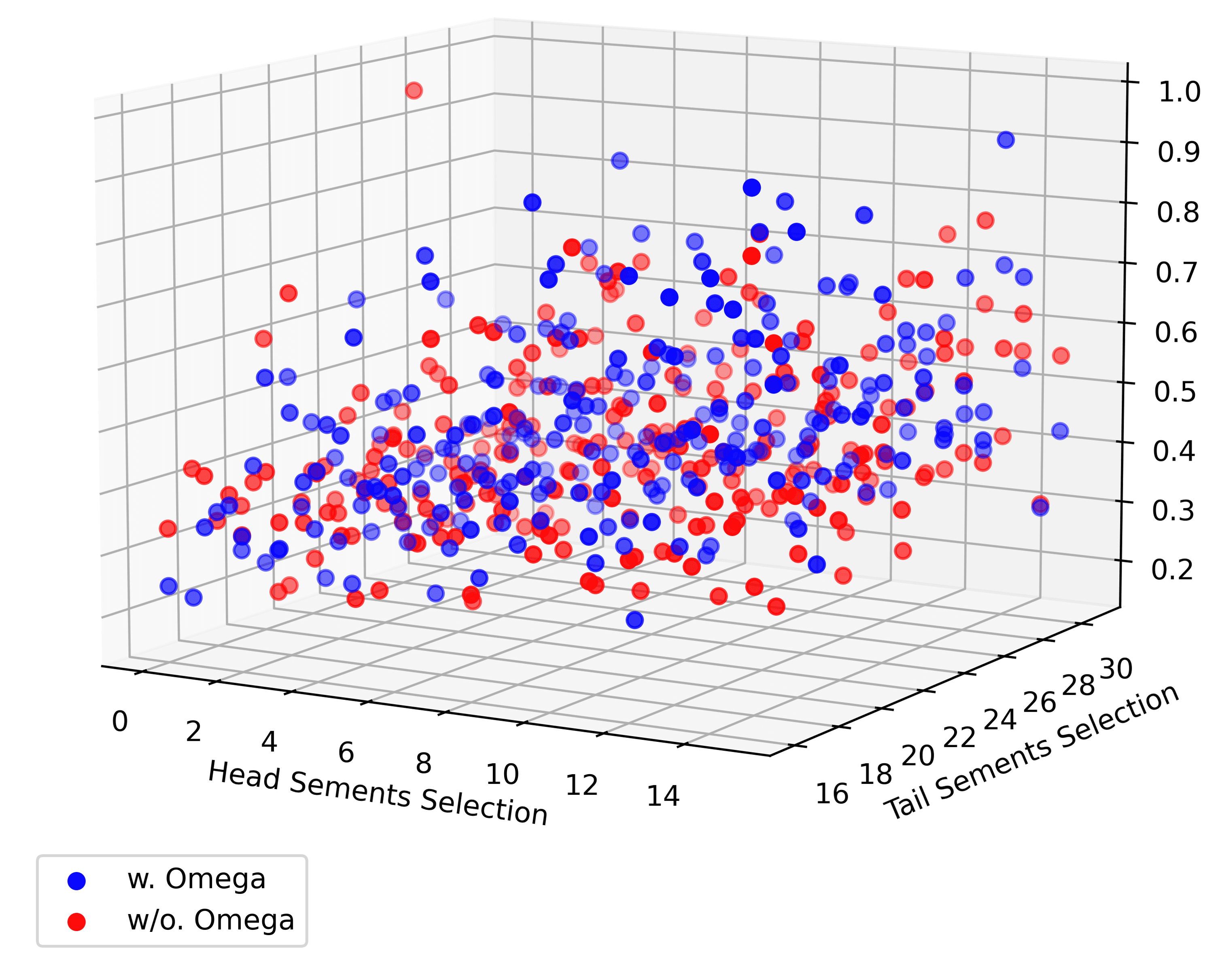} 
    \caption{3D scatter plot comparing the PFID performance with and without Omega, based on the configurations with head segment and tail segment selection.} 
    \label{fig7}
\end{figure}

To validate the effectiveness of re-privatization, we conduct the ablation experiment of $\omega$. Figure~\ref{fig7} demonstrates that selecting omega for weighted operations leads to an overall improvement in LLM performance, indicating the effectiveness of this weighting strategy.

\subsection{Translation Error Test via GPT4 Evaluation}

In order to get more detailed evaluation between our methods and pipelines, We follows GEMBA-MQM \cite{kocmi-federmann-2023-gemba} to evaluate our methods from different translation aspects. We are using \texttt{gpt-4-0125-preview} as the judge, following the prompt shown in the Appendix~\ref{appendix1} . 
Figure~\ref{fig3} showcases our results, which illustrates the significant difference in translation errors between the local and the eavesdropper, especially in critical and major levels. More granular error categorization results can be found in the Appendix~\ref{appendix2}.

\begin{figure}[h]
% % https://colab.research.google.com/drive/1EG82reDnLoigGXugnwAZsoBVfbi9gu87?usp=sharing
\small
\includegraphics[width=0.4\textwidth]{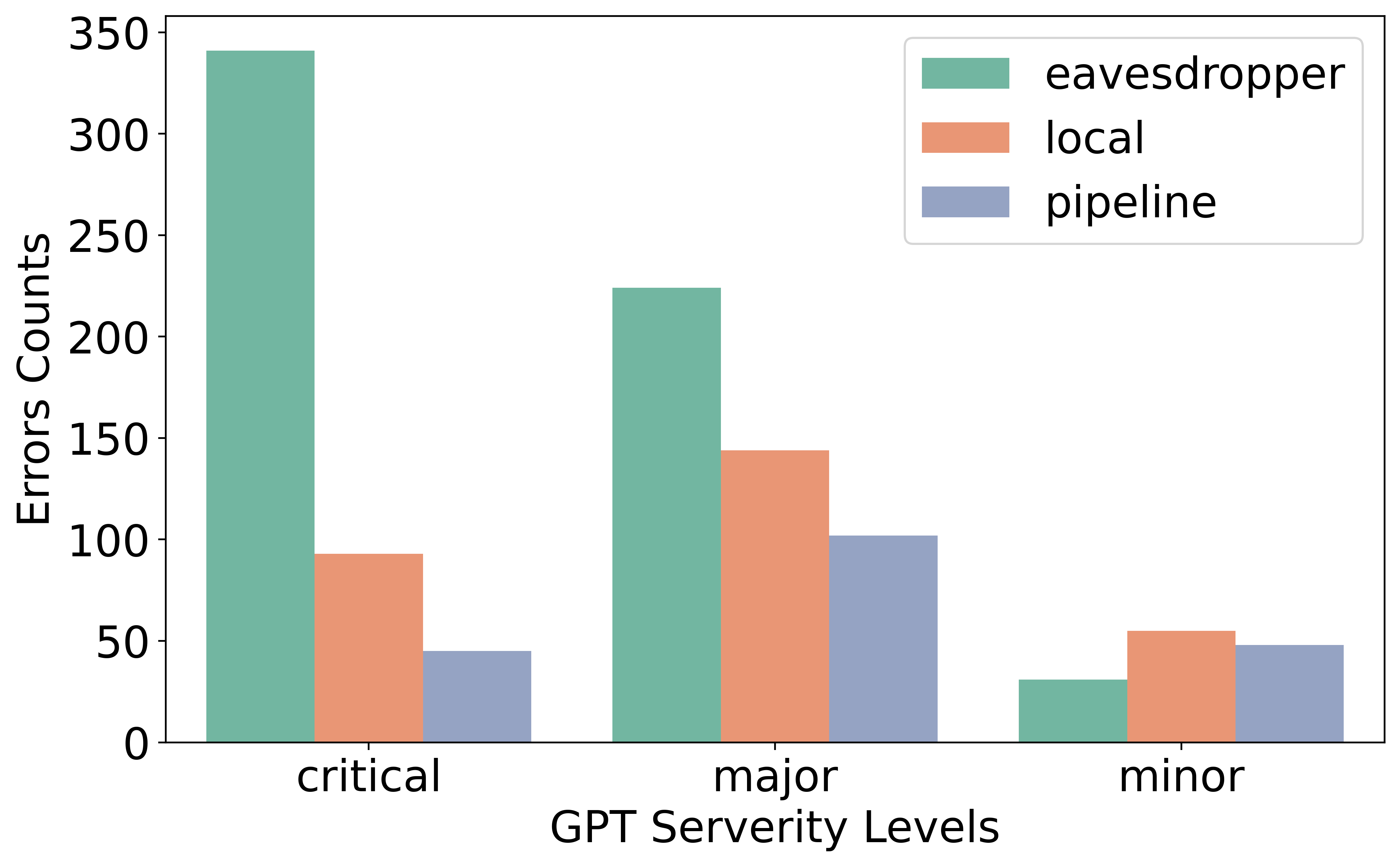}
\centering
\caption{Comparison of error counts in critical, major, and minor levels for pipeline, local, eavesdropper using Vicuna model on partial WMT23 xx-en dataset, the lower the better.}
\label{fig3}
\end{figure}

\subsection{Case Study}
As presetned in Table~\ref{tab4}, two examples indicate that the eavesdropper makes mistakes when decoding stolen private information, such as numbers and ages, while the local can decode correctly, demonstrating the effectiveness of PFID in protecting privacy. Moreover, upon closely examining the first example in the table, we observed that after decoding the information retained locally due to truncation, referred to as "Remnant", it contained accurate privacy information. This interesting discovery suggests that in LLMs, privacy information can be restored by locally truncating the hidden state, lending some interpretability to this method.

\begin{CJK}{UTF8}{gbsn} 
\begin{table*}[t]
\centering
\small
\caption{Two translation examples using Vicuna for zh-en and uk-en language pairs.}
\begin{tabular}{p{15.5cm}}
\hline
\textbf{Source}: 纳斯达克指数昨天涨了2.68\%。(zh-en)\\
\textbf{Truth Answer}: The Nasdaq index yesterday rose \textcolor{DarkGreen}{2.68\%}. \\ 
\hline
\textbf{Local}: The Nasdadaq index rose \textcolor{DarkGreen}{2.68\%} yesterday.\\
\textbf{Eavesdropper}: Nasdadaq index rose rose  ..\textcolor{red}{6\%}. \\
\textbf{Remnant}: daq, is, \textcolor{DarkGreen}{2.68\%}. \\
\hline
\hline
\textbf{Source}: зараз Марії 38 років Вона є хорошою мамою та гарним прикладом для своїх двох синів та однієї донечки. (uk-en)\\
\textbf{Truth Answer}: now Maria is \textcolor{DarkGreen}{38} She is a good mother and a good role model for her two sons and a daughter. \\
\hline
\textbf{Local}: Currently, Maria is \textcolor{DarkGreen}{38} years old. She is a good mother and a good example for her two sons and one daughter.\\
\textbf{Eavesdropper}: Currently Mar Mariaia is \textcolor{red}{33} years old. She is a a mother to a good example for her two sons and one daughter. \\
\hline

\end{tabular}
\label{tab4}
\end{table*}
\end{CJK}

\section{Analysis}

\subsection{Comparative analysis between SVD and noise methods in PFID}
One major method for protecting privacy is differential privacy \cite{Behnia_2022}. As shown in Table \ref{tab6}, we compared experiments involving adding noise only to the hidden state, performing SVD decomposition on the hidden state only, and the combined effects of both approaches. The results revealed that the SVD decomposition used in PFID plays a primary role in protecting privacy. On the other hand, adding noise, as in differential privacy, acts as an indiscriminate form of disruption, making the sentences produced by both local and eavesdroppers less fluent and harder to understand. However, as demonstrated by the examples in Table~\ref{tab7}, this method does not erase critical information, making it challenging to achieve genuine privacy protection.

\begin{table}[]
\centering
\small
\caption{Performance comparison of local and eavesdropper in PFID using SVD, noise application, and their combination.}
\begin{tabular}{llrr}
\hline
                       &             & \multicolumn{1}{l}{Local} & \multicolumn{1}{l}{Eavesdropper} \\
\cline{3-4}
\multirow{3}{*}{COMET} & SVD         & 0.742                     & 0.541                        \\
                       & Noise       & 0.709                     & 0.646                        \\
                       & SVD + Noise & 0.725                     & 0.494                        \\

\hline

\multirow{3}{*}{BLEU}  & SVD         & 15.47                     & 8.2                          \\
                       & Noise       & 12.89                     & 11.67                        \\
                       & SVD + Noise & 12.71                     & 6.48                        \\
\hline
\end{tabular}
\label{tab6}
\end{table}

\begin{CJK}{UTF8}{gbsn} 
\begin{table}[h]
\centering
\small
\caption{Example of noise methods in PFID on zh-en translations.}
\begin{tabular}{p{6.8cm}}
\hline
\textbf{Source}: 天眼查显示，郑永刚持有杉杉控股有限公司的总股权比例是40.1 \%。(zh-en)\\
\textbf{Truth Answer}: According to the SkyEye search, Zheng Yonggang holds \textcolor{DarkGreen}{40.1 \%} of the total equity of Sugo Holdings Co. \\ 
\hline
\textbf{Local}: The display of the sky shows that Zhe Yan holds \textcolor{DarkGreen}{40.1\%} of the shares of Jia Jian Holding Limited.\\
\textbf{Eavesdropper}: Acc display of heaven sky shows that Zhey holds a \textcolor{DarkGreen}{40.1\%} of the equ of Shiajia Holding Co. \\
\hline
\end{tabular}
\label{tab7}
\end{table}
\end{CJK}

\subsection{Framework Effectiveness Analysis via Model Pre-training}
From the perspective of pre-training of LLMs, the effectiveness of PFID in protecting privacy may because during pre-training, these LLMs primarily learn the structure of sentences. Thus, larger eigenvalues correspond to these structural features. Personal information, such as numbers, names, and locations, is learned to a lesser extent, resulting in relatively smaller eigenvalues for these features. By employing SVD decomposition and discarding components with smaller eigenvalues, we can achieve the goal of protecting privacy.

\subsection{Optimal Hyperparameters Selection Analysis}
\begin{figure}[ht]
    \centering
    \begin{subfigure}[t]{0.23\textwidth} 
        \includegraphics[width=\textwidth]{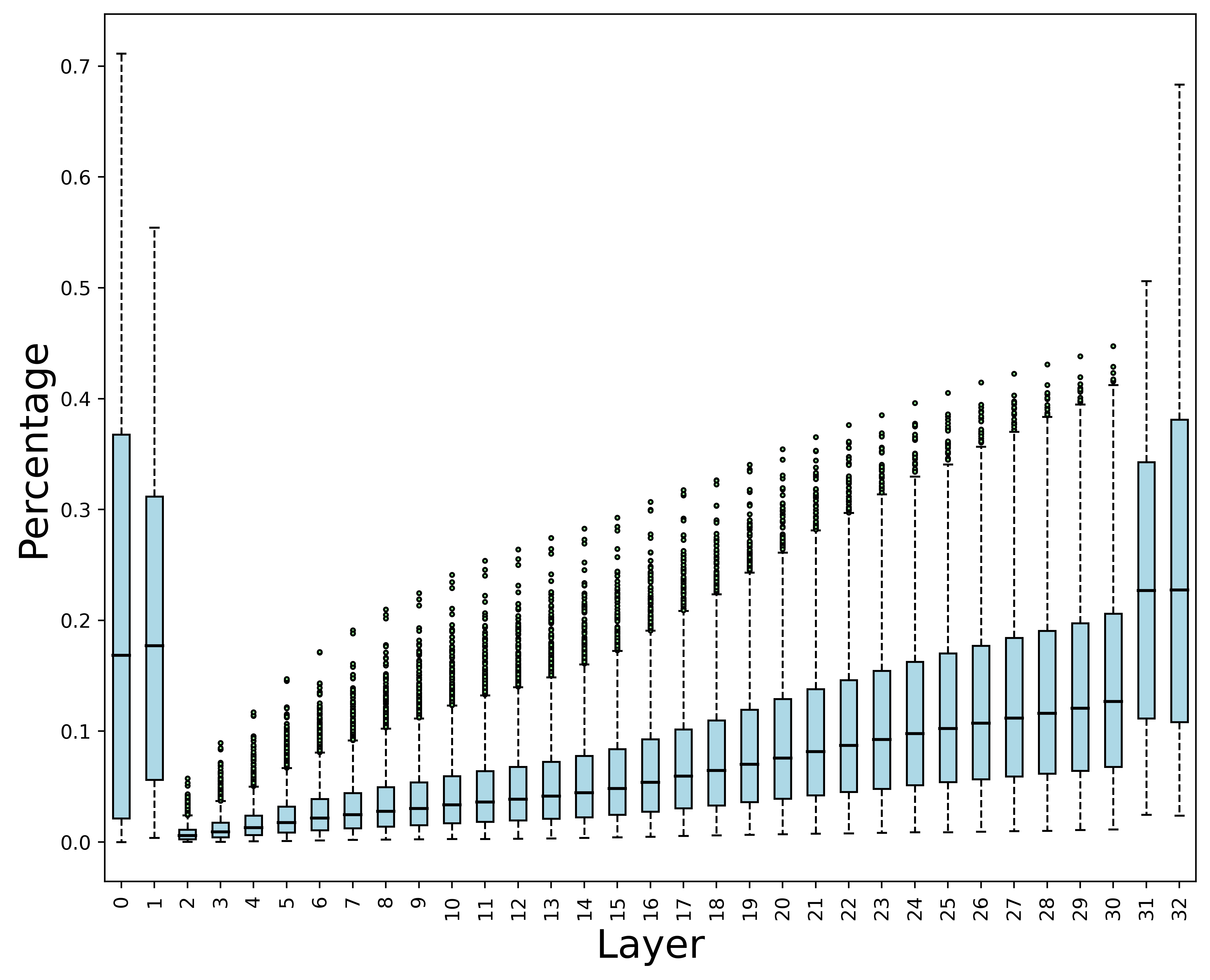} 
        \caption{Cumulative percent of singular values takes up by tail 70 singular values.}
        \label{fig6:sub1}
    \end{subfigure}
    \hfill 
    \begin{subfigure}[t]{0.23\textwidth} 
        \includegraphics[width=\textwidth]{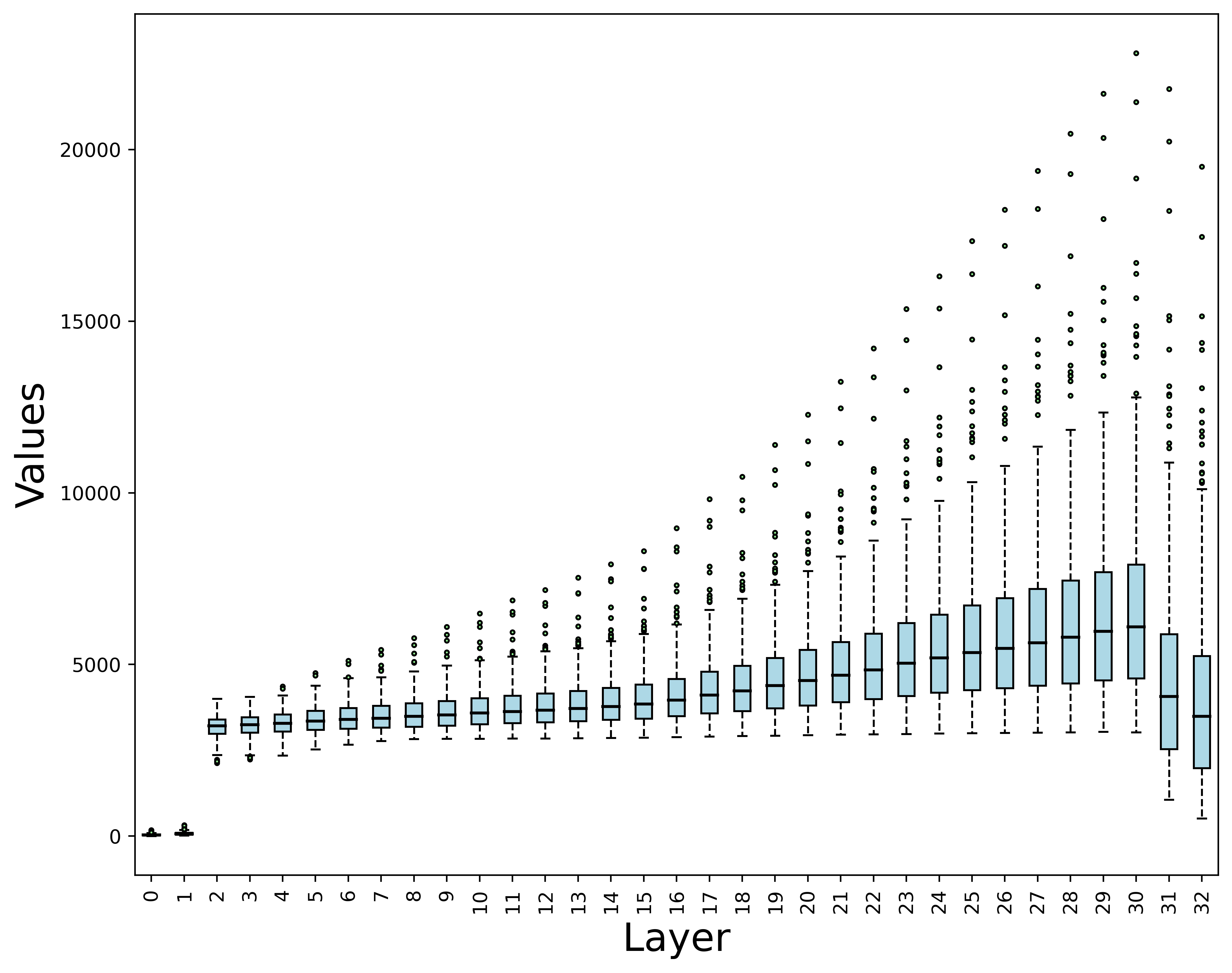}
        \caption{Nuclear norm by layers.}
        \label{fig6:sub2}
    \end{subfigure}
    \caption{Singular values analysis in the Vicuna.}
    \label{fig6}
\end{figure}

As demonstrated in Figure~\ref{fig6:sub1}, the index of the 70\% singular value in LLMs increases with layer depth, indicating a more dispersed distribution of singular values. This suggests that segmenting the model later results in less information capture through SVD. Conversely, Figure~\ref{fig6:sub2} shows increasing singular values, highlighting the growing distinctiveness of information in deeper layers. Early segmentation could omit crucial information. Therefore, mid-model segmentation is identified as optimal.

For the Phead and Ptail, we find that a stronger model allows for a higher truncation ratio with the PFID framework, enhancing privacy protection during transmission between local and public servers. Specifically, with Vicuna at a 0.8 ratio, outputs for both locals and eavesdroppers are disrupted, while Zephyr allows a ratio of 0.9, demonstrating the potential for improved efficiency and security as LLMs evolve.

% As illustrated in Figure~\ref{fig5:sub1}, the index of the 70\% singular value in LLMs tends to increase with the depth of the layers, indicating that the distribution of singular values becomes more dispersed. Therefore, performing model segmentation towards the latter part of the model results in less information being extracted through SVD decomposition. Conversely, as shown in Figure~\ref{fig5:sub2}, the singular values in LLMs increase, suggesting that the importance of information becomes more distinct as the model differentiates further. Consequently, segmenting the model at the very beginning could lead to the loss of some vital information. Hence, conducting model segmentation in the middle is optimal.

\section{Conclusion and Future Work}
In conclusion, our PFID framework demonstrates an efficient method for preserving user privacy in the use of LLMs. By segmenting the model and applying data truncation techniques, we manage to protect sensitive user data from potential breaches while maintaining the high performance and functionality of LLMs. Future work will focus on exploring the framework's applicability to other domains beyond machine translation.

\section*{Limitations}

\noindent
\textbf{Bigger model evaluation}. We have verified our framework effectiveness on three different open-source models, but for even larger model, hidden state's process flow might change, different split dynamic could emerge. \\
\noindent
\textbf{Limitation of input forms}. Our method is designed for auto-regressive prompt input, other non-auto regressive methods might not fit.

\bibliography{anthology,custom}
\bibliographystyle{acl_natbib}

\newpage
\onecolumn
\appendix

\section{Prompt for GPT4 in the translation error test}
\label{appendix1}

\begin{tcolorbox}[title={Machine Translation Quality Assessment Task},fonttitle=\bfseries]
(System) You are an annotator for the quality of machine translation. Your task is to identify errors and assess the quality of the translation.

(user) \textit{\{source\_language\}} source:\\
\{source-segment\}

\textit{\{target\_language\}} translation:\\
\{target-segment\}

Based on the source segment and machine translation surrounded with triple backticks, identify error types in the translation and classify them. The categories of errors are: accuracy (addition, mistranslation, omission, untranslated text), fluency (character encoding, grammar, inconsistency, punctuation, register, spelling), locale convention (currency, date, name, telephone, or time format) style (awkward), terminology (inappropriate for context, inconsistent use), non-translation, other, or no-error.

Each error is classified as one of three categories: critical, major, and minor. Critical errors inhibit comprehension of the text. Major errors disrupt the flow, but what the text is trying to say is still understandable. Minor errors are technically errors, but do not disrupt the flow or hinder comprehension.

(assistant) \textit{\{observed error classes\}}
\end{tcolorbox}

\section{More granular error categorization results}
\label{appendix2}

\begin{figure}[h]
% https://colab.research.google.com/drive/1EG82reDnLoigGXugnwAZsoBVfbi9gu87?usp=sharing
  \centering
  \includegraphics[width=1\textwidth]{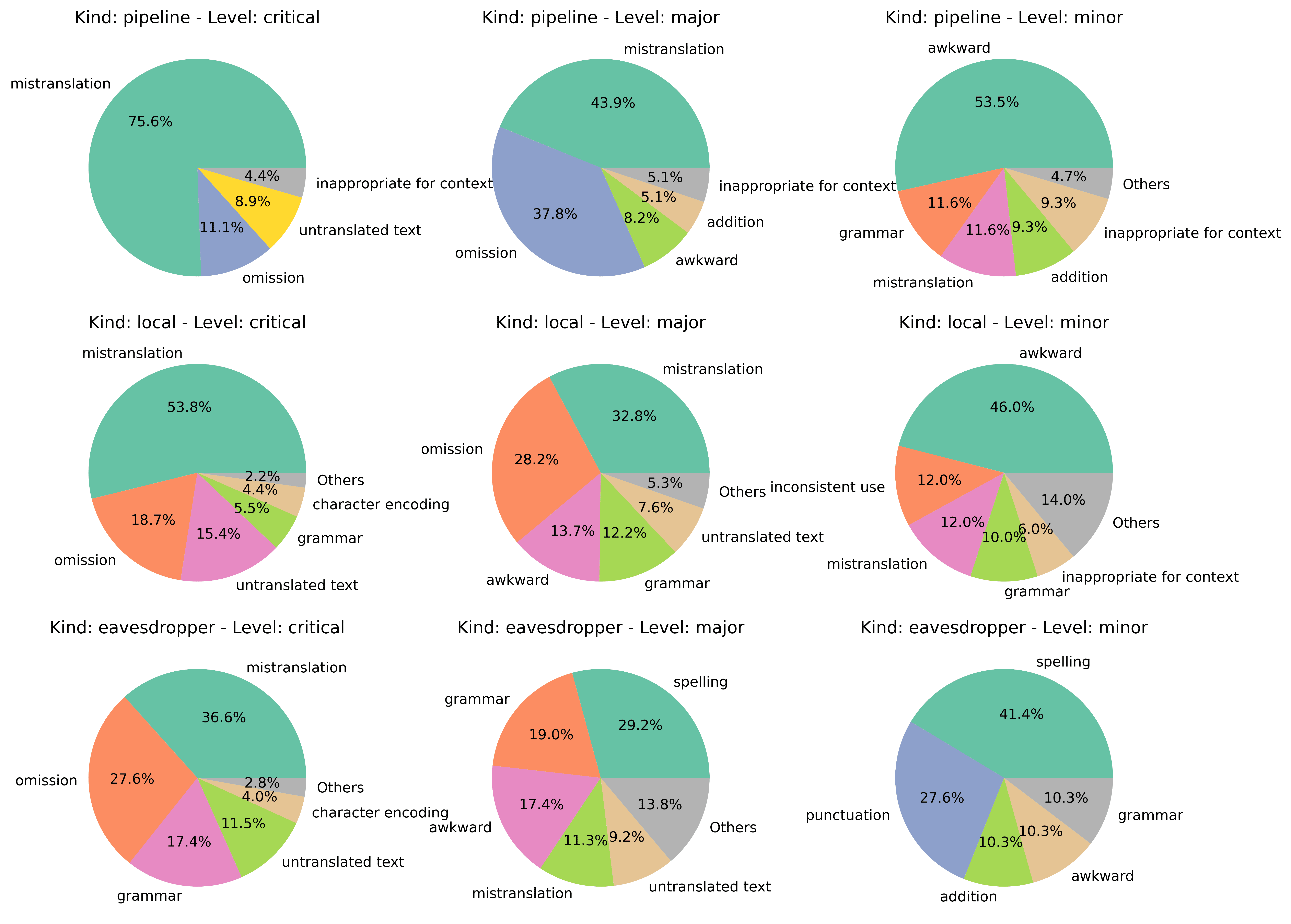}
  \caption{The pie plot with more granular error categorization in critical, major, and minor levels for pipeline, local, eavesdropper using Vicuna model on partial WMT23 xx-en dataset.}
  \label{pieplot-attacker}
\end{figure}

\end{document}